\documentclass{article} 
\usepackage{iclr2022_conference,times}
\usepackage{textcomp}

\usepackage{amsmath,amsfonts,bm}

\def\eqref#1{equation~\ref{#1}}

\def\1{\bm{1}}

\DeclareMathAlphabet{\mathsfit}{\encodingdefault}{\sfdefault}{m}{sl}
\SetMathAlphabet{\mathsfit}{bold}{\encodingdefault}{\sfdefault}{bx}{n}

\usepackage{subcaption}
\usepackage{graphicx}
\usepackage{booktabs}
\usepackage[normalem]{ulem}
\usepackage{multirow}
\usepackage{natbib}
\usepackage{hyperref}
\usepackage{siunitx}
\usepackage{url}
\usepackage{tikz}

\usepackage{titlesec}
\titlespacing*{\section}{0pt}{1.1\baselineskip}{\baselineskip}
\title{NormFormer: Improved Transformer Pretraining with Extra Normalization}

\author{Sam Shleifer\hspace{1em}Jason Weston\hspace{1em}Myle Ott \\
\hspace{5em}Facebook AI Research\thanks{
Jason implemented residual scaling and helped with writing. Myle helped with writing and hardware issues. Thanks to Tim Dettmers for giving us early access to the Adam8Bit Optimizer, and to Naman Goyal, Xian Li, Susan Zhang, Zoe Shleifer and Ofir Press for valuable comments. Correspondence to \url{sshleifer@fb.com}. 
}
 }

\hypersetup{
  colorlinks,
  citecolor=black,
  linkcolor=black,
  urlcolor=blue}

\def\checkmark{\tikz\fill[scale=0.4](0,.35) -- (.25,0) -- (1,.7) -- (.25,.15) -- cycle;} 

\iclrfinalcopy
\begin{document}
\begin{center}
\maketitle
\end{center}

\begin{abstract}
During pretraining, the Pre-LayerNorm transformer suffers from a gradient magnitude mismatch: gradients at early layers are much larger than at later layers. These issues can be alleviated by our proposed NormFormer architecture, which adds three normalization operations to each layer: a Layer Norm after self attention, head-wise scaling of self-attention outputs, and a Layer Norm after the first fully connected layer.
The extra operations incur negligible compute cost (+0.4\% parameter increase), but improve pretraining perplexity and downstream task performance for both causal and masked language models ranging from 125 Million to 2.7 Billion parameters.
For example, adding NormFormer on top of our strongest 1.3B parameter baseline can reach equal perplexity 24\% faster, or converge 0.27 perplexity better in the same compute budget. This model reaches GPT3-Large (1.3B) zero shot performance 60\% faster.
For masked language modeling, NormFormer improves fine-tuned GLUE performance by 1.9\% on average. Code to train NormFormer models is available in \href{https://github.com/pytorch/fairseq/tree/main/examples/normformer}{fairseq}.
\end{abstract}

\section{Introduction}

\begin{figure}[t]
\begin{center}
\includegraphics[scale=0.25]{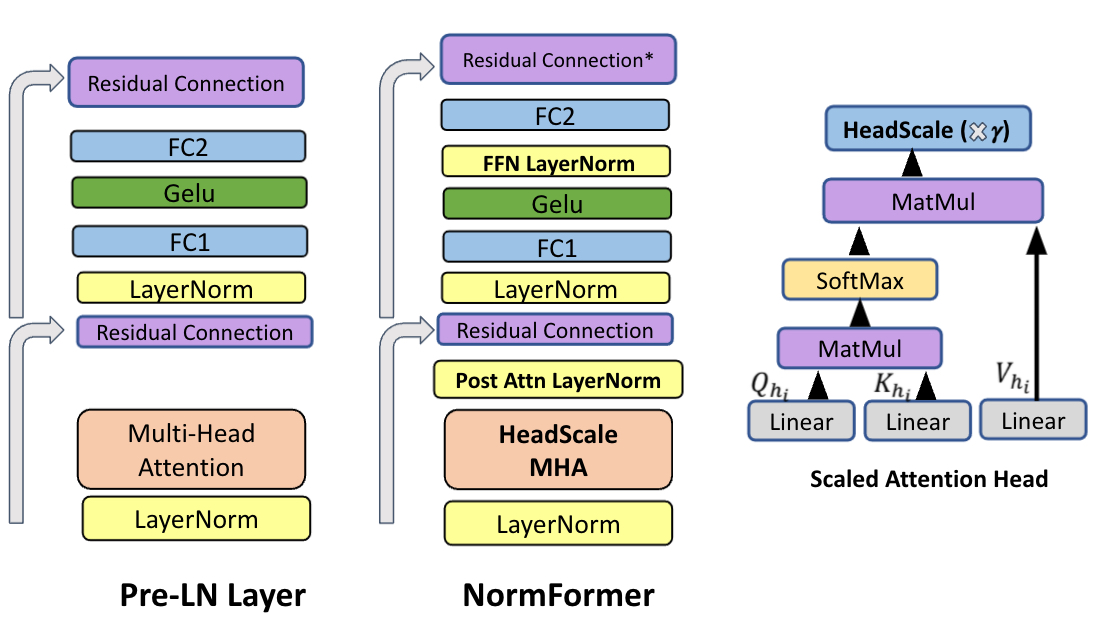}
\caption{Left: a baseline Pre-LayerNorm transformer layer. Center: \texttt{NormFormer}, with the three proposed additions in bold. Right: a single attention head with our proposed \texttt{HeadScale} operation applied prior to the output projection with trainable parameters $\gamma_i$. * When applied, residual scaling impacts the second residual connection in each layer.}
\label{fig:arch}
\end{center}
\end{figure}

The original transformer architecture~\citep{vaswani2017attention} applies Layer Normalization~\citep{ba2016layer} after each sublayer's residual connection (``Post-LN") in order to reduce the variance of the inputs to the following sublayer, i.e.:
\[
\mathrm{PostLN}(x) = \mathrm{LayerNorm}(x + \mathrm{Sublayer}(x)),
\]
with
\begin{align*}
\mathrm{LayerNorm}(x) &= \frac{x - E[x]}{\sqrt{Var[x] + \epsilon}}\cdot \gamma + \beta,
\label{ln_eqn}
\end{align*}
where $\gamma$ and $\beta$ are trainable parameters, and $\epsilon$ is a small constant.
Recent work has observed that Post-LN transformers tend to have 
larger magnitude gradients in later layers compared to earlier layers~\citep{xiong2020layer} and has advocated moving the LayerNorm operation to the beginning of each sublayer (``Pre-LN"; see Figure~\ref{fig:arch}, left), i.e.:
\[
\mathrm{PreLN}(x) = x + \mathrm{Sublayer}(\mathrm{LayerNorm}(x)).
\]
In practice Pre-LN transformers can be trained with larger learning rates, shorter learning rate warmup and often yield improved performance compared to Post-LN transformers~\citep{xiong2020layer}, so most
recent, large pretrained language models tend to use Pre-LN transformers~\citep{baevski2018adaptive, radford2019language, raffel2020exploring, brown2020gpt3, J1WhitePaper}.

In this work we show that, while Pre-LN improves stability over Post-LN, it has the opposite side effect: gradients at earlier layers tend to be larger than gradients at later layers. We propose \texttt{NormFormer}, which alleviates the gradient magnitude mismatch by adding 3 normalization operations to each layer (see Figure~\ref{fig:arch}, middle). These operations reduce gradients to early layers and increase gradients to later layers, bringing their magnitudes closer together.


Compared to compute-matched, well-tuned Pre-LN baselines, \texttt{NormFormer} models reach target pretraining perplexities faster and achieve better pretraining perplexities and downstream task performance.

The rest of this paper is organized as follows: Section~\ref{sec:approach} describes the proposed modifications, Section~\ref{sec:results} shows pretraining and downstream task performance for fully trained \texttt{NormFormer} models against well-tuned, compute-matched baselines. Section~\ref{sec:mismatch} shows the gradient mismatch introduced by Pre-LN and how \texttt{NormFormer} alleviates it. Section~\ref{sec:res-scale} analyzes residual scaling, a related technique proposed to stabilize Post-LN architectures~\citep{xiong2020layer,zhu2021gradinit}. Section~\ref{sec:ablations} shows that removing any of the added operations degrades performance and that \texttt{NormFormer} improves over the baseline at a wide range of hyperparameter configurations.
\section{Approach}
\label{sec:approach}
\subsection{NormFormer}
\texttt{NormFormer} includes three modifications to the Pre-LN transformer: First, we apply head-wise scaling inside the attention module and add two additional LayerNorm operations: one after the attention module and a second after the first fully connected layer.
The modifications introduce a small number of additional learnable parameters, which provide a cost-effective way for each layer to change the magnitude of its features, and therefore the magnitude of the gradients to subsequent components.
The changes are visualized in Figure~\ref{fig:arch} and described below.

\paragraph{Scaling Attention Heads}

The standard multi-head attention operation is defined as:
\begin{align*}
&\mathrm{MultiHeadAttention}(Q,K,V) = \mathrm{Concat}(\mathrm{h}_1,\dots,\mathrm{h}_n) W^O \label{eqn:multihead_att}\\ 
&\mathrm{h}_i = \mathrm{Attention}(QW^Q_i, KW^K_i, VW^V_i) \\
&\mathrm{Attention}(Q,K,V) = \mathrm{softmax}\left(\frac{QK^T}{\sqrt{d_k}}\right)V ,
\end{align*}
where $n$ is the number of heads, $i$ is the attention head index, $d_k$ is the dimensionality of the keys and $W^O,W^Q_i,W^K_i,W^V_i$ are learned projection matrices for the output, query, key and value, respectively.

We propose scaling the output of each attention head via learned scalar coefficients $\gamma_i$:
\begin{align*}
\mathrm{HeadScaleMHA}(Q,K,V) = \mathrm{Concat}(\gamma_1 \mathrm{h}_1,\dots,\gamma_n \mathrm{h}_n) W^O
\end{align*}
where $\gamma$ are learnable parameters initialized to 1.


\paragraph{Additional Layer Normalization and Putting it All Together}

In the Pre-LN transformer each layer $l$ modifies an input $x_l$ as follows:
\begin{align*}
x_{l+1}^{\texttt{PreLN}} &= \mathrm{FFN}(\mathrm{MHA}(x_l)) \\
\\
\mathrm{where\qquad} \mathrm{MHA}(x) &= x + \mathrm{MultiHeadAttention}(\mathrm{LN}(x),\mathrm{LN}(x),\mathrm{LN}(x)) \\
\mathrm{FFN}(x) &= x + \sigma(\mathrm{LN}(x)W_1 + b_1) W_2 + b_2 \\
\mathrm{LN}(x) &= \mathrm{LayerNorm}(x)
\end{align*}
In this work $\sigma$ is the GELU non-linear activation introduced in \citet{hendrycks2016gelu}.

Our overall method, \texttt{NormFormer}, instead modifies each input $x_l$ as: 
\begin{align*}
x_{l+1}^{\texttt{NormFormer}} &= \mathrm{NormFFN}(\mathrm{NormScaledMHA}(x_l)) \\
\\
\mathrm{where\qquad} \mathrm{NormScaledMHA}(x) &= x + \mathrm{\textbf{LN}}(\mathrm{\bf{HeadScaleMHA}}(\mathrm{LN}(x),\mathrm{LN}(x),\mathrm{LN}(x)))\\
\mathrm{NormFFN}(x) &= ~\bm{x} + \mathrm{\textbf{LN}}(\sigma(\mathrm{LN}(x)W_1 + b_1)) W_2 + b_2
\end{align*}
where bolded operations are newly introduced. 

\subsection{Experiments}\label{sec:experiments}

\paragraph{Causal Language Models}
We pretrain causal LMs (CLM) that roughly match the ``Small" (125M parameter), ``Medium" (355M), ``Large" (1.3B) and ``XL" (2.7B) sizes from \citet{brown2020gpt3}.

Our model architecture differs from \citet{brown2020gpt3} in two ways: (1) we use only dense attention, while they alternate between dense and locally banded sparse attention; (2) we train our models with sinusoidal positional embeddings, following Shortformer~\citep{press2020shortformer}, since early experiments found this to produce comparable results with fewer learned parameters.

We train the baseline models for 300 billion tokens.
We train \texttt{NormFormer} models for an equivalent number of GPU hours, which typically results in 2-6\% fewer steps and tokens due to the additional overhead of the normalization operations.

\begin{table}[t]
\begin{center}
\begin{tabular}{@{}cccc@{}}
\toprule
Model Size & GPT-3 Paper & Baseline & NormFormer \\ \midrule
125M       & 6e-4    & 3e-3 & 3e-3   \\
355M       & 3e-4    & 1e-3 & 1e-3   \\
1.3B       & 2e-4    & 6e-4 & 6e-4   \\\bottomrule
\end{tabular}
\caption{Searching for learning rates on our dataset results in higher values than reported in~\citet{brown2020gpt3}, providing stronger baselines to compare to our NormFormer architecture.}
\label{tab:lr}
\end{center}
\end{table}

On our dataset, we find that the learning rates proposed in GPT-3 are suboptimally low.\footnote{The difference in optimal learning rates may be due partly to architectural differences between our baseline and GPT-3 (e.g., not using locally banded sparse attention).} For both baseline and NormFormer at each size besides 2.7B, we tune the learning rate by training models for 50,000 steps and selecting the best performing learning rate among: $\{\num{1e-4}, \num{6e-4}, \num{3e-4}, \num{6e-4}, \num{1e-3}, \num{3e-3}\}$. The learning rates we obtained from this process, shown in Table~\ref{tab:lr}, are 3-5 times larger than those used in the GPT-3 paper. Additionally, we have verified that the baseline and NormFormer both perform worse at the full training budget with the GPT-3 learning rates than with the higher learning rates. Other hyperparameters do not differ from GPT-3.\footnote{See Table 2.1 in~\citet{brown2020gpt3}.}

\paragraph{Residual Scaling}

Standard Post-LN transformers simply sum the previous output (residual) with the new output. Recent work attempts to stabilize Post-LN architectures by weighting the residual connection for each layer~\citep{zhu2021gradinit, liu2020understanding}.
We thus experiment with scaling the residual in each embedding dimension via learned scalar coefficients  $(\lambda_{resid})_i$:
\[
\mathrm{ResScale(x)} =  \lambda_{resid}  \circ x+ \mathrm{Sublayer}(\mathrm{LayerNorm}(x))
\]
where $\circ$ is elementwise multiplication, and $\lambda_{resid}$ are learned parameters initialized to 1.

While this can be applied at any normalization layer, we find it it most effective for normalizing the feedforward network (FFN) submodule for the smaller sized language models.
In this setting, 
\begin{align*}
\mathrm{NormFFN}(x) &= {\bm{\lambda}_{\bm{resid}}~{\bm{\circ}}~\bm{x}} + \mathrm{\textbf{LN}}(\sigma(\mathrm{LN}(x)W_1 + b_1)) W_2 + b_2
\end{align*}
For 1.3B parameter models and larger, scaling residuals hurts performance (see discussion in Section~\ref{sec:res-scale}), so \texttt{ResScale} is not used in our 1.3B and 2.7B CLM results.

\paragraph{Large scale experiments}
We also train three large-scale models with 2.7B parameters.
Our first baseline is a replicated version of GPT-3-2.7B with GELU activations, the published learning rate (1.6e-4) and the same number of training steps and tokens (286K steps; 300B tokens). This model slightly exceeds the reference zero shot performance~\citep{brown2020gpt3}.
Next, we train two variants of GPT3-2.7B with $Relu^{2}$ activations~\citep{so2021primer}, but use slightly fewer training steps (20\% less) for compute efficiency.
The first of these uses the baseline learning rate (1.6e-4) and the second uses \texttt{NormFormer-2.7B} with a higher learning rate of 6e-4.
We note that training baseline 2.7B CLMs (i.e., without \texttt{NormFormer} modifications) with a higher 6e-4 learning rate diverged and failed to train.
However, as opposed to the smaller architectures, we did not exhaustively tune the learning rate, so it is possible that an intermediate value would perform better.

\paragraph{Zero Shot Evaluation}
In addition to validation perplexity, we evaluate CLMs on a subset of the tasks that GPT3 evaluated on in a zero-shot setting~\citep{ brown2020gpt3}, with the same prompts. We select WinoGrande~\citep{sakaguchi2020winogrande}, StoryCloze~\citep{mostafazadeh-etal-2016-corpus}, OpenBookQA~\citep{mihaylov-etal-2018-suit}, HellaSwag~\citep{zellers-etal-2019-hellaswag} and PIQA~\citep{bisk2020piqa} because GPT3 showed strong performance on these tasks at small scale, as well as consistently improving performance with scale.

\paragraph{Masked Language Models (MLM)}
We adopt the RoBERTa-base, Pre-LN architecture and hyperparameters used in~\citet{liu2019roberta}.
For the baseline, we pretrain for 2 million batches of 1 million tokens, about $\frac{1}{4}$ of the training budget of the original \texttt{roberta-base}.
NormFormer runs through 1.92 million batches in the same amount of time.

\paragraph{Fine-Tuning}
We fine-tune both the baseline MLM and NormFormer with learning rates
$\num{1e-5}, \num{1e-4}, \num{3e-4}, \num{1e-3}, \num{3e-3}, \num{6e-3}$ and report the best performance on the validation set for each GLUE task~\citep{wang2019glue}, following \citet{liu2019roberta}.
Other fine-tuning hyperparameters match those used for \texttt{roberta-base} in \citet{liu2019roberta}.

\paragraph{Pretraining data}
We pretrain all models on a collection of English language text including the English portion of the CC100 corpus~\citep{conneau2020unsupervised} as well as the data from \citet{liu2019roberta}, consisting of BookCorpus~\citep{zhu2015bookcorpus}, English Wikipedia and filtered subsets of Common Crawl.
We encode our data with the byte-level Byte Pair Encoding (BPE) vocabulary from \citet{liu2019roberta}, originally introduced in \citet{radford2019language}.
The combined dataset contains around 450GB of uncompressed text and 110B BPE tokens.
We hold out 40M BPE tokens from this data as a validation set on which we report pretraining perplexities.


\paragraph{Implementation details}
We train our causal and masked language models in \texttt{fairseq}~\citep{ott2019fairseq, paszke2019pytorch}.
Although NormFormer introduces fewer than 0.07\% additional parameters, it slows individual training updates and increases memory usage between 2\% (2.7B model) to 6\% (125M model) due to the FFN LNs.
Accordingly, we compare NormFormer to baseline models trained for an equal amount of GPU time, i.e., controlling for compute rather than the number of training updates.
Finally, we note that the \texttt{HeadScale} operation can be moved outside the self attention module to allow the use of the very efficient pytorch \texttt{F.multihead\_attention}. This change reduces overhead without noticeable performance degradation.

\section{Results}
\label{sec:results}
\begin{table}
\centering
\resizebox{\textwidth}{!}{
\begin{tabular}{cccccc|c|ccccc|c}
\toprule
{} & $|\theta|$ &      LR & $Relu^{2}$ & $\lambda_{resid}$ & Steps &    PPL &    HS &    PI &    WG &    SC &    OB &   Avg \\
\midrule
Random Baseline            &        - &     - &        - &          - &     - &      - &  25.0 &  50.0 &  50.0 &  50.0 &  25.0 &  40.0 \\\midrule
GPT3-125M (paper)          &      124.4 &    6e-4 &          - &            - &    572K &      - &  33.7 &  64.6 &  52.0 &  63.3 &  35.6 &  49.8 \\
GPT3-125M (replicated)     &      124.4 &    6e-4 &          - &            - &    572K &  21.11 &  33.7 &  66.5 &  52.2 &  66.1 &  35.4 &  50.8 \\
GPT3-125M (High LR)        &      124.4 &    3e-3 &          - &            - &    572K &  21.09 &  35.3 &  67.5 &  50.5 &  66.3 &  35.0 &  50.9 \\
NormFormer-125M           &      124.5 &    3e-3 &          - &            - &    540K &  20.34 &  34.9 &  67.1 &  52.3 &  66.3 &  38.0 &  51.7 \\
NormFormer-125M            &      124.5 &    3e-3 &          - &    \checkmark &    539K &  \textbf{20.11} &  34.9 &  65.9 &  53.4 &  67.5 &  40.0 &  \textbf{52.3} \\\midrule
GPT3-355M (paper)          &      354.7 &    3e-4 &          - &            - &    572K &      - &  43.6 &  70.2 &  52.1 &  68.5 &  43.2 &  55.5 \\
GPT3-355M (replicated)     &      354.7 &    3e-4 &          - &            - &    572K &  15.41 &  46.1 &  70.8 &  54.6 &  71.1 &  41.2 &  56.8 \\
GPT3-355M (High LR)        &      354.7 &    1e-3 &          - &            - &    572K &  14.85 &  48.4 &  71.7 &  53.8 &  73.3 &  43.4 &  58.1 \\
NormFormer-355M           &      355.0 &    1e-3 &          - &            - &    552K &  14.54 &  49.7 &  71.8 &  56.0 &  73.8 &  43.6 &  59.0 \\
NormFormer-355M            &      355.0 &    1e-3 &          - &    \checkmark &    550K &  \textbf{14.52} &  49.7 &  72.0 &  56.7 &  73.2 &  43.8 &  \textbf{59.1} \\\midrule
GPT3-1.3B (paper)          &     1313.5 &    2e-4 &          - &            - &    286K &      - &  54.7 &  75.1 &  58.0 &  73.4 &  46.8 &  61.6 \\
GPT3-1.3B (replicated)     &     1313.5 &    2e-4 &          - &            - &    286K &  12.56 &  58.5 &  74.6 &  58.1 &  76.8 &  49.4 &  63.5 \\
GPT3-1.3B (High LR)        &     1313.5 &    6e-4 &          - &            - &    286K &  12.21 &  57.5 &  74.3 &  59.3 &  76.3 &  50.8 &  63.6 \\
NormFormer-1.3B            &     1314.0 &    6e-4 &          - &            - &    275K &  \textbf{11.94} &  60.5 &  74.5 &  60.1 &  77.5 &  50.8 &  \textbf{64.7} \\\midrule
GPT3-2.7B (paper)          &     2648.7 &  1.6e-4 &          - &            - &    286K &      - &  62.8 &  75.6 &  62.3 &  77.2 &  53.0 &  66.2 \\
GPT3-2.7B (replicated)     &     2648.7 &  1.6e-4 &          - &            - &    286K &  10.92 &  65.9 &  76.6 &  61.4 &  78.2 &  49.6 &  66.3 \\
NormFormer-2.7B       &     2649.5 &    6e-4 &  \checkmark &            - &    277K &  \textbf{10.55} &  68.1 &  78.1 &  64.4 &  79.4 &  53.4 &  \textbf{68.7} \\\midrule
GPT3-2.7B-Relu             &     2648.7 &  1.6e-4 &  \checkmark &            - &    230K &  10.99 &  65.9 &  76.1 &  63.2 &  79.3 &  49.4 &  66.8 \\
GPT3-2.7B-Relu             &     2648.7 &  6e-4 &  \checkmark &            - & 28K &   \multicolumn{7}{c}{diverged} \\
NormFormer-2.7B &     2649.5 &    6e-4 &  \checkmark &            - &    222K &  \textbf{10.73} &  67.4 &  77.2 &  64.4 &  78.9 &  52.6 &  \textbf{68.1} \\
\bottomrule
\end{tabular}
}
\caption{Zero-Shot Accuracy for Causal LMs for the following tasks: \texttt{HS}: HellaSwag, \texttt{PI}: PIQA, \texttt{WG}: WinoGrande, \texttt{SC}: StoryCloze, \texttt{OB}: OpenBookQA. PPL is validation perplexity during pretraining. \textit{GPT-3 (paper)} results taken from~\citet{brown2020gpt3}. Horizontal lines group compute-matched runs. \emph{High LR} corresponds to using a larger learning rate than reported in \citet{brown2020gpt3}.
$\lambda_{resid}$ indicates whether residual scaling was used. $\lambda_{resid}$ did not help at 1.3B scale, as shown in \ref{fig:clm_pt}, but that run is not compute matched so it is not included here. Model size ($|\theta|$) is reported in millions of parameters.
}
\label{tab:lm_zero_shot}
\end{table}
\begin{figure}[t]
\centering
\includegraphics[width=0.85\textwidth]{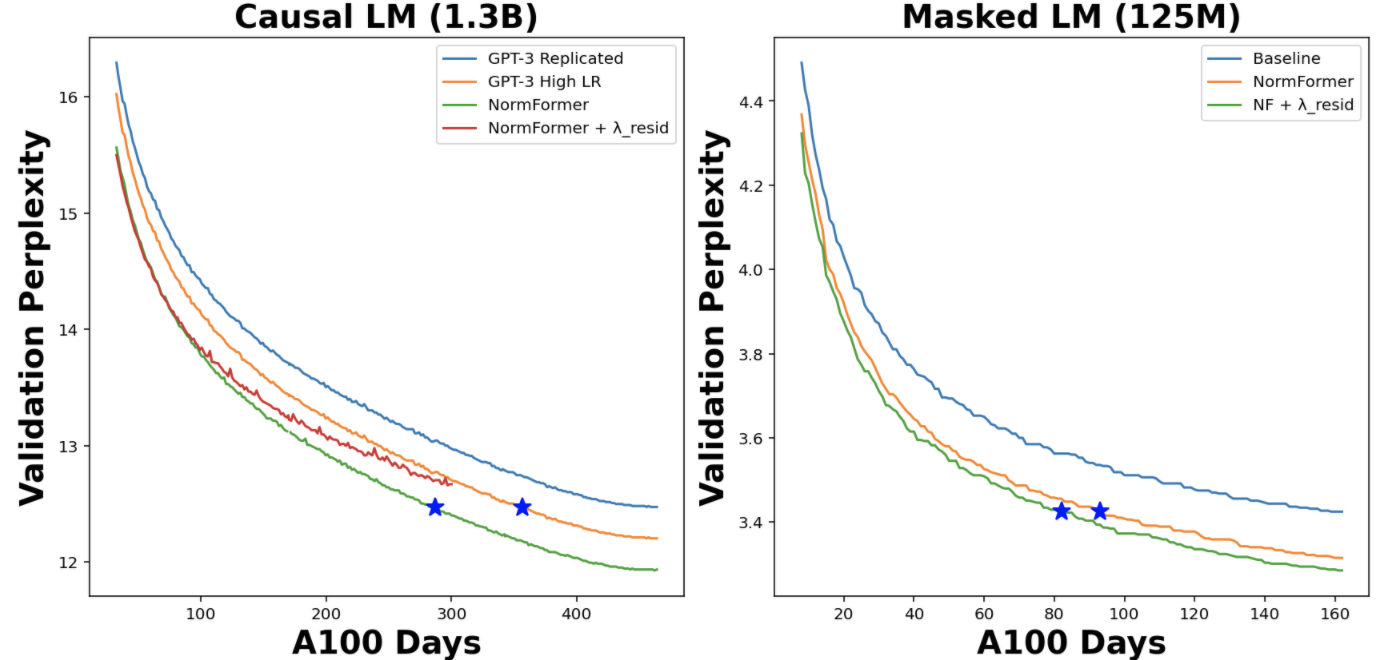}
\caption{Pretraining perplexity on held-out validation data for Causal and Masked Language Models as a function of training compute (GPU days). The blue stars show the point where a model matches the baseline's lowest perplexity. 
}
\label{fig:clm_pt}
\end{figure}

We report pretraining perplexities for CLMs and MLMs as a function of training wall-time (GPU days) in Figure~\ref{fig:clm_pt}.
We observe that NormFormer trains significantly faster and achieves better validation perplexities for a given training compute budget.
The blue stars mark the first validation step where NormFormer matches the baseline's lowest perplexity and shows that NormFormer matches Pre-LN models while needing only 60\% and 57\% as much compute for CLM and MLM models, respectively.
This is particularly impressive since NormFormer models take 2-6\% longer for each training step and thus see less data than Pre-LN models in this comparison. The left side blue line in Figure~\ref{fig:clm_pt} shows the failed attempt to add \texttt{ResScale} to \texttt{NormFormer-1.3B}.

We observe a similar trend on downstream tasks.
In Table~\ref{tab:lm_zero_shot} we report zero shot accuracy for causal LMs using the tasks and prompts from~\citet{brown2020gpt3}.
NormFormer outperforms GPT-3 at all sizes.
The gains from \texttt{Normformer} extra parameters operations outpace the gains from normal scaling laws. Changing the hidden dimension of a 125M parameter model from 768 to 780, for example, results in a 127 million parameter model that is only 0.08 perplexity better than the baseline whereas \texttt{NormFormer-125M} adds only 100,000 parameters and is 0.83 perplexity better than the baseline.

\begin{table}[t]
\centering
\resizebox{\textwidth}{!}{
\begin{tabular}{lcc|c|ccccccc|c}
\toprule
{} &  Model Size &  $\lambda_{resid}$ &   PPL &  CoLA &  MNLI &  MRPC &  QNLI &   QQP &   RTE &  SST-2 &    Avg \\\midrule
Baseline     &      125.42 &    - &  3.42 &  74.3 &  85.9 &  84.6 &  91.6 &  90.7 &  66.4 &   92.9 &  83.77 \\
NormFormer  &      125.50 &    - &  3.31 &  \textbf{82.6} &  \textbf{86.3 }&  \textbf{86.0} &  \textbf{91.9} &  \textbf{91.3} & \textbf{ 67.9} &   93.8 &  \textbf{85.69} \\
NormFormer   &      125.51 &     \checkmark &  \textbf{3.29} &  80.9 &  86.2 &  85.3 &  91.5 &  91.2 &  62.8 &   \textbf{94.2} &  84.59 \\
\bottomrule
\end{tabular}
}
\caption{Masked LM: Pretraining validation perplexity (PPL) and fine-tuned performance on GLUE tasks for Pre-LN and NormFormer models. Note that models are trained for an equal amount of compute, which is less than the publicly-released \texttt{roberta-base} models.}
\label{tab:roberta_ft2}
\end{table}

For MLM models, we report fine-tuned accuracy on GLUE in Table~\ref{tab:roberta_ft2}.
We again find that NormFormer MLM models outperform their Pre-LN counterparts on every task (rows 1 vs 2).
Adding \texttt{ResScale} improves improves pre-training performance marginally (3.29 valid PPL vs 3.31), but the gains to do not translate to finetuned performance.

\section{Analysis}

\label{sec:mismatch}

\subsection{Analysis of gradient norms by layer}
\label{sec:grad-norm}

\begin{figure}[t]
\centering
\includegraphics[width=\linewidth]{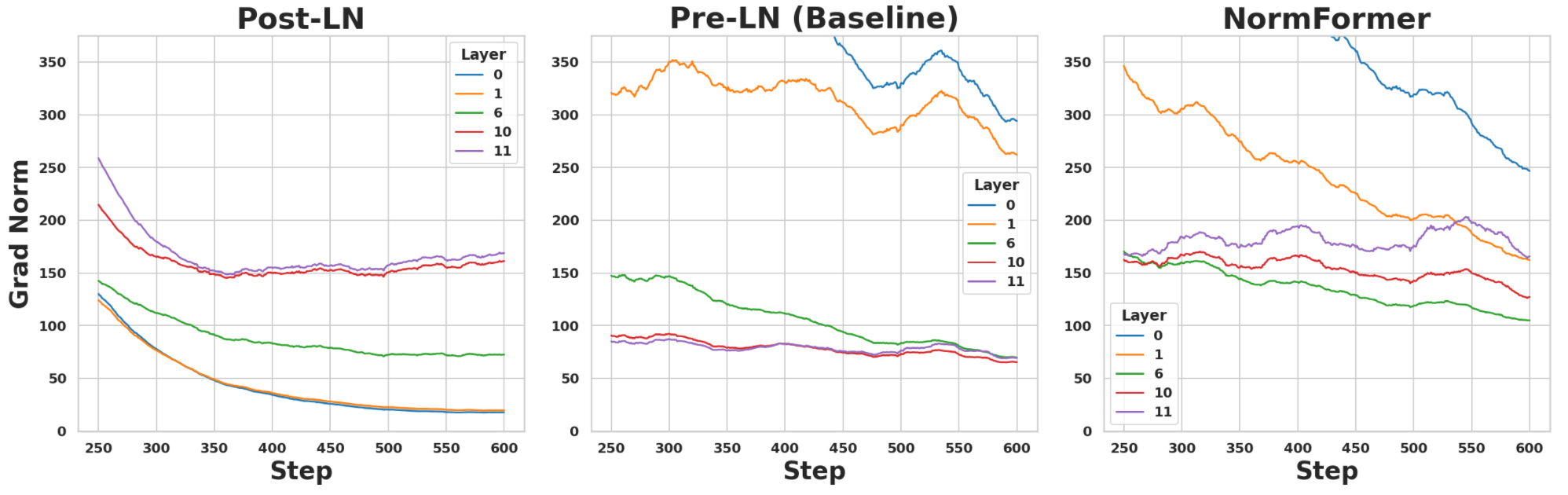}
\caption{Average L1 norm of gradients to the second fully connected weight for layers 0,1,6,10 and 11, early in training. }
\label{fig:gnorm_fc2}
\end{figure}
\begin{figure}[t]
\centering
\includegraphics[scale=0.2]{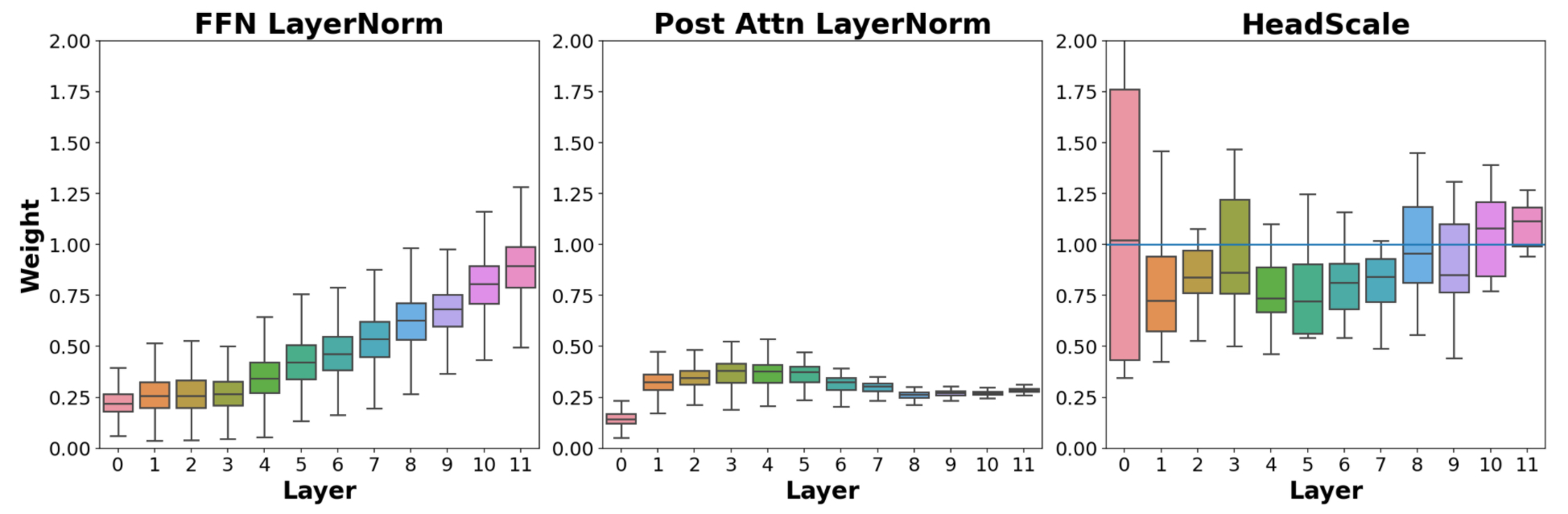}
\caption{Distribution of learned scaling parameters in three of the added operations. For FFN LN, earlier layers receive downscaled inputs, keeping their gradients in the same range as the gradients of later layers. This plot is discussed in detail in Section~\ref{sec:mismatch}.}
\label{fig:ffn_layernorm}
\end{figure}
\begin{figure}[ht]
\centering
\includegraphics[scale=0.25]{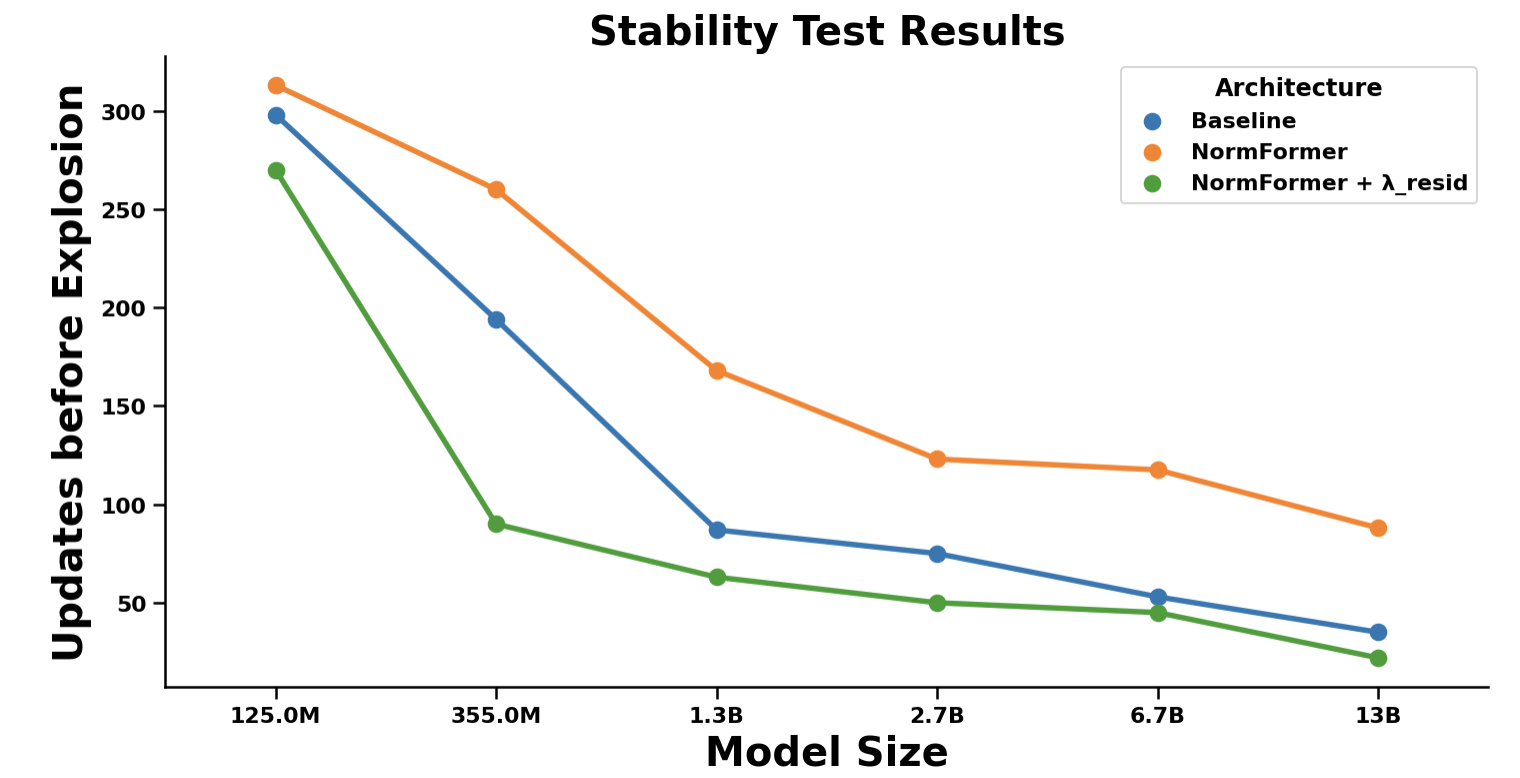}
\caption{LR Stability Test: learning rate starts from 0 and linearly increases by \texttt{5e-5} at each training step until training destabilizes. NormFormer reaches a higher learning rate before destabilizing. Each data point is the median of 3 runs with a different random seed.}
\label{fig:stability-results}
\end{figure}

We begin by examining the magnitude of the gradients at different layers for Post-LN, Pre-LN and NormFormer models, since large magnitude differences in gradients across layers can destabilize training, particularly when training in mixed precision~\citep{micikevicius2018mixed}.
Figure~\ref{fig:gnorm_fc2} shows the average L1 norm of the gradients to the second fully connected weight in various layers for a 12 layer, 125M parameter CLM model at the beginning of training.
As reported in past work~\citep{xiong2020layer}, we observe that the gradients to later layers in Post-LN models are much larger than for earlier layers, and that the gradients to early layers quickly vanish in the early stages of training.
Pre-LN models have the opposite behavior, with early layers instead receiving significantly larger gradients than later layers.
\texttt{NormFormer} brings the average gradient norms closer together for different layers in the network.

In Figure~\ref{fig:ffn_layernorm} we present the distribution of scaling parameters learned by \texttt{NormFormer} models.
For the FFN LN, the $\gamma$ parameters are smaller for earlier layers, reducing the magnitude of the inputs to early fully connected parameters, thereby decreasing the magnitude of their gradients.
The post attention LN, in the middle of Figure~\ref{fig:ffn_layernorm}, all layers have $\gamma$ coefficients below 1, indicating downscaling.\footnote{The downscaling is also apparent in Figure~\ref{fig:gnorm_all} in the Appendix, which plots the change in grad norm for each operation at each layer. It shows that adding extra normalization reduces the gradient norm for all attention parameters at every layer. Only FFN parameters at later layers, have increased gradient norms.} The \texttt{HeadScale} $\gamma$ parameters, shown in the rightmost plot in Figure~\ref{fig:ffn_layernorm} vary more than the others, and have no relationship with depth in the network. We interpret this as evidence that the \texttt{HeadScale} parameters dynamically increase the importance of well initialized attention heads, as suggested in \citet{chen2021earlybert}.

One result of reducing the gradient mismatch, besides better perplexities and downstream task performance, is the ability to train stably with larger learning rates.
To measure the stability of an architecture, we train it on a learning rate schedule with a very large peak learning rate, so that the learning rate increases a little each step until the loss explodes. Figure~\ref{fig:stability-results} shows that NormFormer models can survive for more updates in this environment than the baseline. For the baseline 125M model (the left most blue dot), the loss eventually explodes, with the activations from multiplying the query and key features at layer 0 overflowing the FP16 range. The down scaling of the attention outputs allows NormFormer to avoid this issue and remain stable with larger learning rates.  Figure~\ref{fig:stability-results} also shows that $\lambda_{resid}$ reduces the stability improvement at all sizes.

\subsection{Residual Scaling}
\label{sec:res-scale}
By comparing adjacent NormFormer-125M and NormFormer-355M rows in Table~\ref{tab:lm_zero_shot} we can see that adding \texttt{ResScale} to \texttt{NormFormer} improves perplexity and zero shot performance for small scale CLMs. For 125M parameter MLM, \texttt{ResScale} improves pre-training perplexity marginally, but hurts fine-tuned performance.  At 1.3 billion parameter scale, however, adding \texttt{ResScale} to \texttt{NormFormer} does not improve performance (Figure~\ref{fig:clm_pt}). Although it’s not included in our tables, we find that \texttt{ResScale} without NormFormer is stronger than the baseline at small scale, but not large scale. This suggests that the negative result is caused by scale, rather than interaction with \texttt{NormFormer}.

\begin{figure}[ht]
\centering
\includegraphics[scale=0.2]{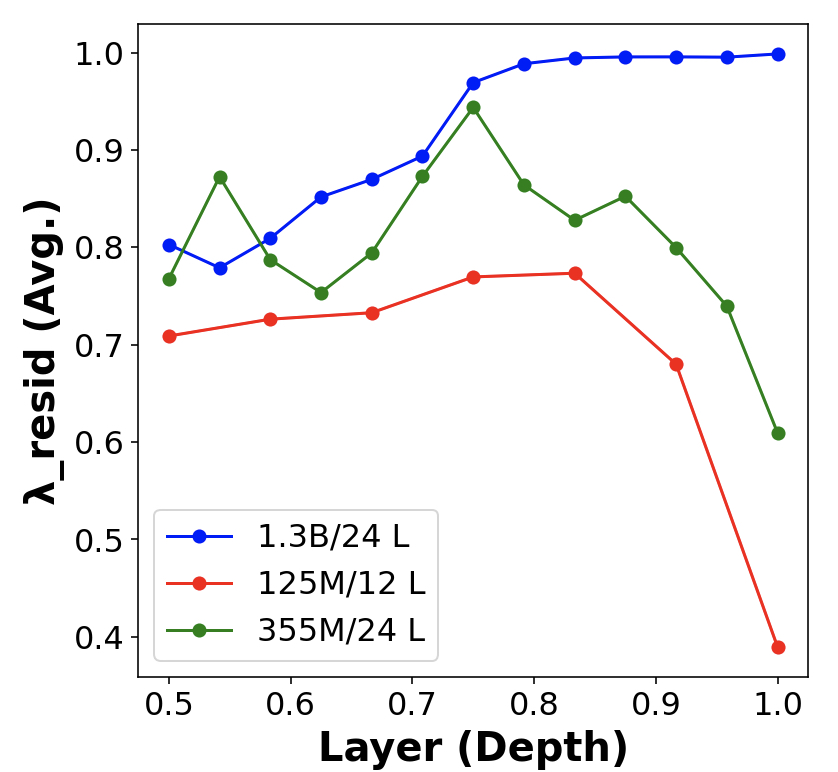}
\caption{$\lambda_{resid}$ weights at each layer of different sized CLMs in the \texttt{NormFormer+$\lambda_{resid}$} setting. Depth is layer number / total layers.}
\label{fig:resid-coeff}
\end{figure}

Figure~\ref{fig:resid-coeff} shows the Avg. $\lambda_{resid}$ weights at each layer of different sized CLMs. We can see that at 125M and 355M parameters, the weights in the later layers are lower, indicating down weighting of the residual connection, whereas at the largest scale, 1.3B, the weights are larger deeper into the network.

Adding the $\lambda_{resid}$ parameters to the other (earlier) residual connection in each layer, or using a scalar instead of a vector for each $\lambda_{resid}$, does not fix the large scale issue, but hurts small scale performance marginally.

\section{Ablations}
\label{sec:ablations}
This section provides evidence that removing any of our additions to the transformer block degrades performance on language modeling tasks, and that our additions improve language modeling performance across a wide range of hyperparameter settings. Experiments use 125M parameter CLMs, and are run with the default hyperparameters given in Table~\ref{tab:ablation_hparams} in the appendix for 470 V100 Hours (100,000 updates for the baseline) unless otherwise mentioned.

\paragraph{Removing any of the added operations hurts performance}

Table~\ref{tab:more_ln} shows that none of the four introduced operations can be removed without degrading performance.
Rows 2-5 remove each operation one at a time. In all cases perplexity increases, with the removal of \texttt{HeadScale} being the most damaging and the removal of the Post-Attn LN being the least damaging.
In Row 6 (\texttt{+ 3 More LN}) we try to introduce more normalization inside self attention, applying LN to the query, key and value features in addition to our 3 other operations, for a total of 6 new operations. In this setting, every other parameterized operation inside the transformer layer is an LN.  We find that this does not change perplexities at a fixed number of updates, but reduces training speed by another 5\%.  This result suggests that there is not much upside to adding even more normalization on top of \texttt{NormFormer}.

\begin{table}[t]
\begin{center}
\begin{tabular}{@{}lc@{}}
\toprule
Architecture   & Valid PPL \\ \midrule
NormFormer+ResScale      & \textbf{15.88}     \\
\hspace{8pt}- Post-Attn LN & 15.92     \\
\hspace{8pt}- FFN LN       & 16.14     \\
\hspace{8pt}- Head Scale   & 16.22     \\ 
\hspace{8pt}- Res Scale    & 16.20     \\ 
\hspace{8pt}+ 3 More LN         & 15.88     \\ \midrule
Baseline       & 16.37     \\ \bottomrule
\end{tabular}
\caption{125M parameter Language Modeling Validation perplexities after 470 V100 Hours of pretraining. Removing any of our proposed additions degrades performance (Rows 2-5). Adding more normalization inside the Multi Headed Attention (Row 6) does not impact perplexity at a fixed number of updates, but reduces throughput such that the model can only complete 87,500 updates vs. 92,500 for Rows 1-5 and 100,000 for Row 7. Note that these PPL scores are not directly comparable to other tables -- they use a different validation set.}
\label{tab:more_ln}
\end{center}
\end{table}

\paragraph{Other Experiments}
Replacing the FFN LN with the FFNGeGlu proposed in \citet{shazeer2020glu}, which includes scaling but no normalization, degraded performance in our 125M parameter CLM setting, the only place we tried it. We also find that the LN variant proposed in \citet{raffel2020exploring}, which removes the bias and the mean substraction from the normalization, performs equally well to our LN and has fewer trainable parameters, but is about 2x slower than the \texttt{FusedLayerNorm} implementation we use. We therefore do not adopt it.

\citet{ding2021cogview} propose related stabilization strategies for text to image generation tasks with larger models including a downscaled embedding gradient, a layer norm after the final fully connected layer, and the same post-attention LN. We find that, besides the post attention LN, these techniques do not help in our setting.

Table~\ref{tab:hparam_results} in the appendix shows language modeling perplexities for 7 different hyperparameter configurations, separated by horizontal lines.
\texttt{NormFormer} outperforms the baseline in all settings.


\section{Related Work}  
Layer normalization \citep{ba2016layer} is an important component of the transformer architecture.
\citet{xiong2020layer} shows that for Post-LN: gradients are too big for later layers and solves this problem with Pre-LN. We build on the Pre-LN architecture to make it even more stable and efficient.

\citet{press2020improving} proposes an architecture where instead of interleaving attention and feed forward sublayers, the attention all happens first. This increases the number of late FFN parameters, rather than increasing their importance and gradient norm, as our FFN LN does, and does not impact stability.

Our \texttt{HeadScale} operation is related to that used in \citet{chen2021earlybert}, but used differently. Whereas that work prunes attention heads with low $\gamma$ parameters, we use the $\gamma$ parameters to improve pretraining performance.

These approaches are also related to techniques for initializing neural networks: GradInit \citep{zhu2021gradinit} introduces a set of scalars and biases for initialization based on a variance heuristic, and Admin \citep{liu2020understanding} applies a similar heuristic in profiling and initialization stages. These works  also use variants of our \texttt{ResScale} operation, which we find helpful at small scale and harmful at large scale.

Similarly, some other approaches targeted initialization as well, in particular ReZero \citep{bachlechner2020rezero}, FixUp \citep{huang2020improving} and LookLinear \citep{balduzzi2017shattered}. We note that DALL-E \citep{ramesh2021zero} also added a per residual scaling factor (only during backprop).
Our approach, in contrast, only has new learnable parameters without variance heuristics, and has no extra stages or changes in initialization.

\section{Conclusion}
We identify a mismatch in the gradients of Pre-LN transformer weights: earlier layers receive much larger gradients than later layers, while the optimal scaling of residuals is larger at earlier layers than at later layers. 
We propose \texttt{NormFormer}, which alleviates these issues by adding 3 extra operations to each transformer layer. These modifications help the gradient mismatch for fully connected parameters and improve validation perplexity and downstream task performance for both causal and masked language models. None can be removed without degrading performance back towards the baseline, and adding more normalization -- at least of the types we have tried -- does not improve performance.
Since NormFormer primarily addresses the gradient mismatch by increasing the gradients to the last FFN layers while decreasing the gradient magnitudes in other parts of the network, future work could examine whether all 3 operations need to be added to every layer.
Additionally, the small computational overhead associated with NormFormer could be alleviated by fusing the FFN LN with the preceding fully connected layer, with or without the mean centering and bias, which do not appear to improve pretraining perplexity. 
In general, we have shown that adding small numbers of learnable parameters in the right places in our architectures can alleviate certain issues in current state of the art networks. Future work should ascertain if there are additional similarly efficient modifications that can bring gains, while helping us understand current deficiencies further.

\newpage
\section{Appendix}\label{sec:appendix}

\begin{table}[!h]
\begin{center}
\begin{tabular}{@{}cccc@{}}
\toprule
          & Learning Rate & Setting Changes & Valid PPL      \\ \midrule
Baseline  & 0.001         &    -           & 16.80          \\
NormFormer & 0.001         &    -          & \textbf{16.33} \\ \midrule 
Baseline  & 0.003         &    -          & 16.37          \\ 
NormFormer & 0.003         &    -           & \textbf{15.88} \\ \midrule
Baseline  & 0.006         &   -              & 16.58          \\
NormFormer & 0.006         &   -            & \textbf{16.22} \\ \midrule
Baseline  & 0.003         & Longer Warmup          & 16.50          \\
NormFormer & 0.003         & Longer Warmup         & \textbf{16.06} \\ \midrule
Baseline  & 0.003         & GPT3          & 16.29          \\
NormFormer & 0.003         & GPT3          & \textbf{15.88} \\\midrule
Baseline  & 0.003         & Clip Grad Norms at 0.1          & 16.46          \\
NormFormer & 0.003         & Clip Grad Norms at 0.1          & \textbf{16.14} \\ \bottomrule
\end{tabular}
\caption{Longer Warmup: increase LR Warmup to 6,000 steps (from 500). GPT3: increase sequence length to 2048, increase dropout to 0.1, increase training budget to 1,000 V100 hours. Grad Clip: clip gradient norms at 0.1. NormFormer outperforms the baseline in all settings.}
\label{tab:hparam_results}
\end{center}
\end{table}

\paragraph{Wikitext103}
\label{wikitext}

Table~\ref{tab:wikitext} shows that NormFormer can also provide gains on top of a well tuned language model in settings with much less data. We simply add our three operations to the architecture and hyperparameters of \citet{baevski2018adaptive}. Convergence perplexity improves, and we reach the baseline perplexity in 70\% as many steps. In this setting, \texttt{NormFormer} does not improve in the last 30\% of training, which suggests that with more tuning the perplexity gap could be widened.

\begin{table}[htpb]
\centering
\begin{tabular}{ccc}
\toprule
{} & Steps to Final PPL &  PPL \\
\midrule
Baseline   &            100\% &      18.70 \\
NormFormer &            70\% &      18.65 \\
\bottomrule
\end{tabular}
\caption{Wikitext 103 results following \citet{baevski2018adaptive}. \texttt{Steps to Final PPL}: at what percentage of the 280K steps did the model reach 18.70 perplexity.
\texttt{PPL}: Best Perplexity}
\label{tab:wikitext}
\end{table}
\begin{figure}[ht]
\centering
\includegraphics[scale=0.25]{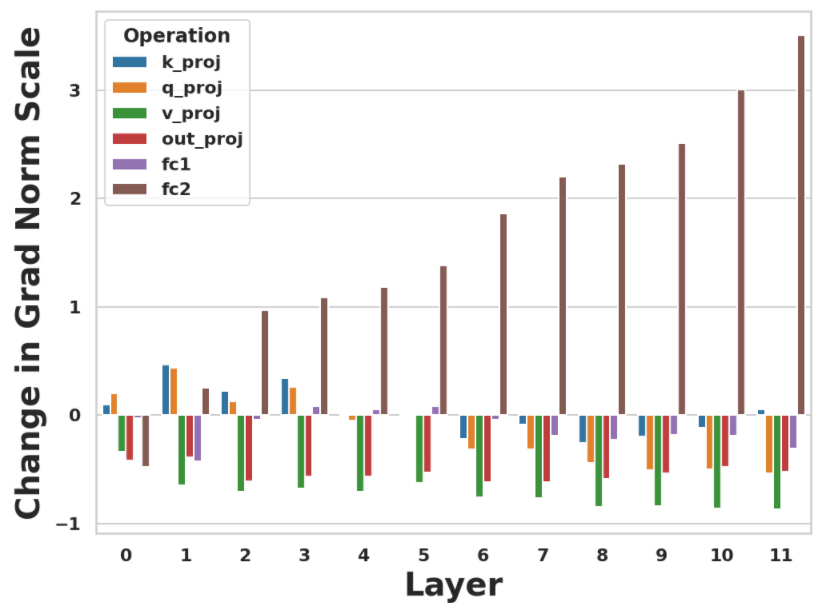}
\caption{Change in grad norm with each operation of NormFormer compared to the baseline. Norms are the average between step 950 and 1000, normalized to control for different losses. 2.0 on the Y axis means the gradient to a parameter is twice as large as the baseline, on average. The NormFormer increases the norm to fully connected parameters in later layers, while reducing the gradient norm to attention parameters at all layers. The results are discussed in detail in Section~\ref{sec:mismatch}.}
\label{fig:gnorm_all}
\end{figure}

\begin{table}[h]
\begin{center}
\small
\begin{tabular}{@{}cc@{}}
\toprule
Learning Rate     & 0.003            \\ 
Batch Size        & 524K Tokens      \\
Parameters        & 124M+            \\
Layers            & 12               \\
Layer Dimension   & 768              \\
Dropout           & 0                \\
LR Warmup Updates & 500              \\
LR Scheduler      & Linear Decay \\
Sequence Length   & 1024             \\
Train Budget      & 470 V100 Hours   \\ \bottomrule
\end{tabular}
\caption{Hyperparameters for ablations in Tables~\ref{tab:more_ln} and~\ref{tab:ablation_hparams}. This train budget allows the baseline model to run for 100,000 updates.}
\label{tab:ablation_hparams}
\end{center}
\end{table}

\newpage
\bibliography{legend}
\bibliographystyle{iclr2022_conference}


\end{document}